# Solving Fourier ptychographic imaging problems via neural network modeling and TensorFlow

SHAOWEI JIANG,[1,3] KAIKAI GUO,[1,3] JUN LIAO,[1,3] AND GUOAN ZHENG[1, 2*]

[1]*Biomedical Engineering, University of Connecticut, Storrs, CT, 06269, USA*
[2]*Electrical and Computer Engineering, University of Connecticut, Storrs, CT, 06269, USA*
[3]*These authors contributed equally to this work*
*guoan.zheng@uconn.edu

**Abstract:** Fourier ptychography is a recently developed imaging approach for large field-of-view and high-resolution microscopy. Here we model the Fourier ptychographic forward imaging process using a convolution neural network (CNN) and recover the complex object information in the network training process. In this approach, the input of the network is the point spread function in the spatial domain or the coherent transfer function in the Fourier domain. The object is treated as 2D learnable weights of a convolution or a multiplication layer. The output of the network is modeled as the loss function we aim to minimize. The batch size of the network corresponds to the number of captured low-resolution images in one forward / backward pass. We use a popular open-source machine learning library, TensorFlow, for setting up the network and conducting the optimization process. We analyze the performance of different learning rates, different solvers, and different batch sizes. It is shown that a large batch size with the Adam optimizer achieves the best performance in general. To accelerate the phase retrieval process, we also discuss a strategy to implement Fourier-magnitude projection using a multiplication neural network model. Since convolution and multiplication are the two most-common operations in imaging modeling, the reported approach may provide a new perspective to examine many coherent and incoherent systems. As a demonstration, we discuss the extensions of the reported networks for modeling single-pixel imaging and structured illumination microscopy (SIM). 4-frame resolution doubling is demonstrated using a neural network for SIM. The link between imaging systems and neural network modeling may enable the use of machine-learning hardware such as neural engine and tensor processing unit for accelerating the image reconstruction process. We have made our implementation code open-source for the broad research community.

## 1. Introduction

Many biomedical applications require imaging with both large field of view and high resolution at the same time. One example is whole slide imaging (WSI) in digital pathology, which converts tissue sections into digital images that can be viewed, managed, and analyzed on computer screens. To this end, Fourier ptychography (FP) is a recently developed coherent imaging approach for achieving both large field of view and high resolution at the same time [1-4]. This approach integrates the concepts of aperture synthesizing [5-11] and phase retrieval [12-20] for recovering the complex object information. In a typical microscopy setting, FP sequentially illuminates the sample with angle-varied plane waves and uses a low numerical aperture (NA) objective lens for image acquisition. Changing the incident angle of the illumination beam results in a shift of the light field's Fourier spectrum at the pupil plane of the objective lens. Therefore, part of the light field that would normally lie outside the pupil aperture can now transmit through the system and be detected by the image sensor. To recover the complex object information, FP iteratively synthesizes the captured intensity images in the Fourier space (aperture synthesizing) and recover the phase information (phase retrieval) at the same time. The final achievable resolution of FP is determined by the synthesized passband at the Fourier space. As such, it is able to use a low-NA objective with a low-magnification factor to produce a high-resolution complex object image, combining the advantages of wide field of view and high resolution at the same time [21-23].

The FP approach is also closely related to the real-space ptychography, which is a lensless phase retrieval technique originally proposed for transmission electron microscopy [14, 24-26]. Real-space ptychography employs a confined beam for sample illumination and records the Fourier diffraction patterns as the sample is mechanically scanned to different positions. FP has a similar operating principle as real-space ptychography but switching the real space and the Fourier space using a lens [1, 22, 27]. The mechanical scanning of the sample in real-space ptychography is replaced by the angle scanning process in FP. Despite the difference in hardware implementation, many algorithm developments of real-space ptychography can be directly applied in FP, including the sub-sampling scheme [28], the coherent-state multiplexing scheme [29, 30], the multi-slice modeling approach [31, 32], and the object-probe recovering scheme [15, 33].

Here we model the Fourier ptychographic forward imaging process using a convolution neural network (CNN) and recover the complex object information in the network training process. In this approach, the input layer of the network is the point spread function (PSF) in the spatial domain or the coherent transfer function (CTF) in the Fourier domain. The object is treated as 2D learnable weights of a convolution or a multiplication layer. The output of the network is modeled as the loss function we aim to minimize. The batch size of the network corresponds to the number of captured low-resolution images in one forward / backward pass. We use a popular open-source machine learning library, TensorFlow [34], for setting up the network and conducting the optimization process. We analyze the performance of different learning rates, different solvers, and different batch sizes. It is shown that a large batch size with the Adam optimizer achieves the best performance in general. To accelerate the phase retrieval process, we also discuss a strategy to implement Fourier-magnitude projection using a multiplication neural network model. Since convolution and multiplication are the two most-common operations in imaging modeling, the reported approach may provide a new perspective to examine many coherent and incoherent systems. As a demonstration, we discuss the extension of the reported networks for modeling single-pixel imaging and structured illumination microscopy. 4-frame resolution doubling is demonstrated using a neural network for structured illumination microscopy. The link between imaging systems and neural network models may enable the use of machine-learning hardware such as neural engine (also known as AI chips) and tensor processing unit [35] for accelerating the image reconstruction process. We have made our implementation code open-source for the broad research community.

This paper is structured as follows: in Section 2, we discuss the forward imaging model for the Fourier ptychographic imaging process and propose a CNN for modeling this process. We then analyze the performance of different learning rates, different solvers, and different batch sizes of the proposed CNN. In Section 3, we discuss a strategy to implement the Fourier-magnitude projection using a multiplication neural network model. In Section 4, we discuss the extension of the reported approach for modeling single-pixel imaging and structured illumination microscopy via CNNs. Finally, we summarize the results and discuss our on-going efforts in Section 5.

## 2. Modelling Fourier ptychography using a convolution neural network

The forward imaging process of FP can be expressed as

$$I_n(x,y) = \left|(O(x,y) \cdot e^{i(k_{xn}x+k_{yn}y)}) * PSF(x,y)\right|^2, \quad (1)$$

where '·' denotes element-wise multiplication, '*' denotes convolution, $O(x,y)$ denotes the complex object, $e^{i(k_{xn}x+k_{yn}y)}$ denotes the $n^{th}$ illumination plane wave with a wave vector $(k_{xn}, k_{yn})$, $PSF(x,y)$ denotes the PSF of the objective lens, and $I_n(x,y)$ denotes the $n^{th}$ intensity measurement by the image sensor. The Fourier transform of $PSF(x,y)$ is the CTF of the objective lens. For diffraction-limited imaging, we have $FT\{PSF(x,y)\} = circ(k_x^2 + k_y^2 < (NA \cdot k_0)^2)$, where $FT$ denotes Fourie transform, $k_0 = 2\pi/\lambda$ and $\lambda$ is the wavelength, $circ$ is the circle function (it is 1 if the condition $k_x^2 + k_y^2 < k_0^2$ is met and 0 otherwise). Equation (1) can be rewritten as

$$I_n(x,y) = \left|O(x,y) * (PSF(x,y) \cdot e^{-i(k_{xn}x+k_{yn}y)})\right|^2 = |O(x,y) * PSF_n(x,y)|^2, \quad (2)$$

where $PSF_n(x,y) = PSF(x,y) \cdot e^{-i(k_{xn}x+k_{yn}y)}$. In most existing machine learning libraries, the learning weights need to be real numbers. As such, we need to expand the complex object and the PSF with the following equations:

$$O(x,y) = O_r(x,y) + i \cdot O_i(x,y), \quad PSF_n(x,y) = PSF_{nr}(x,y) + i \cdot PSF_{ni}(x,y), \quad (3)$$

where subscript 'r' denotes the real part and 'i' denotes the imaginary part. Combining Eqs. (2) and (3), we can express the forward imaging model of FP as

$$I_n(x,y) = (PSF_{nr} * O_r - PSF_{ni} * O_i)^2 + (PSF_{ni} * O_r + PSF_{nr} * O_i)^2, \quad (4)$$

The goal of the Fourier ptychographic imaging process is to recover $O_r$ and $O_i$ based on many measurements $I_n(x,y)$ (n = 1,2,3…). Since $O_r$ and $O_i$ are real number, we can model it as a learnable two-channel filter in a CNN.

The proposed CNN model for the Fourier ptychographic imaging process is shown in Fig. 1. This model contains an input layer for the $n^{th}$ PSF, a convolution layer with two channels for the real and imaginary parts of the object, an activation layer for performing the square operation, an add layer for adding the two inputs, and an output layer for the predicted FP image. For the convolution layer, we can choose different stride value to model the down-sampling effect of the image sensor. In our implementation, we choose a stride value of 4, i.e., the pixel size of the object is 4 times smaller than that of the FP low-resolution measurements.

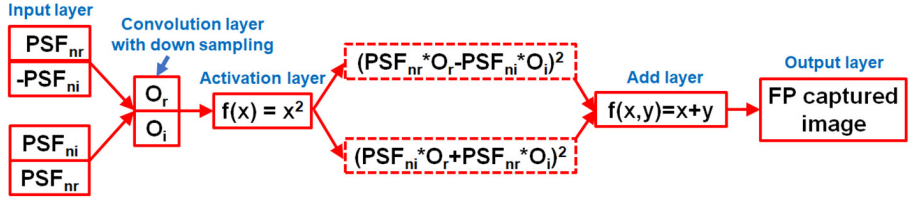

Fig. 1 The CNN model for the Fourier ptychographic imaging process. The input of the network is the n[th] PSF. The object is treated as a two-channel learnable filter of a convolution layer. We use a stride of 4 for the convolution layer, and thus, the pixel size of the output intensity images is 4 times larger than that of the object. The output of the network represents the captured FP image. The optimization process of the network is to minimize Eq. (6). The implementation code is provided in dataset 1.

The training process of this CNN model is to recover the two-channel object ($O_r$, $O_i$) based on all FP captured images $I_n(x, y)$ (n = 1,2,3…). For an initial guess of the two-channel object ($O_r$, $O_i$), the CNN in Fig. 1 outputs a prediction $I_{n\_predict}$ in the forward pass:

$$I_{n\_predict}(x,y) = (PSF_{nr} * O_r - PSF_{ni} * O_i)^2 + (PSF_{ni} * O_r + PSF_{nr} * O_i)^2, \quad (5)$$

In the backward pass, the difference between the prediction and the captured FP image $diff(I_n, I_{n_{predict}})$ is back-propagated to the convolution layer and the two-channel object is updated accordingly. Therefore, the training process of the CNN model can be viewed as a minimization process for the following loss function:

$$loss = diff\left(I_n, I_{n_{predict}}\right) = \sum_{n=i}^{i-1+batchSize} |I_n - I_{n\_predict}|, \quad (6)$$

where L1-norm is used to measure the difference between the prediction and the actual measurement, and 'batchSize' corresponds to the number of images in one forward / backward pass. If the batch size equals to 1, it is stochastic gradient descent with the gradient evaluated by a single image at one forward / backward pass. If the batch size equals to the total number of measurements, it is similar to using Wirtinger derivatives and gradient descent scheme to recover the complex object [36], except that we use L1-norm in Eq. (6) instead of L2-norm (the difference between L1 / L2 norms will be discussed in a later section).

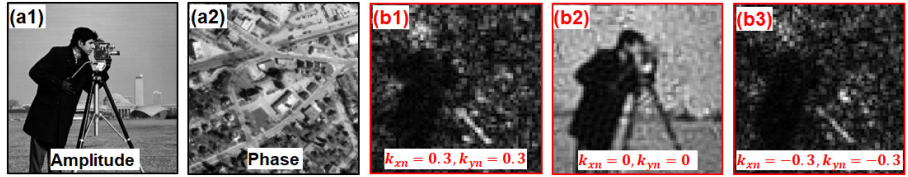

Fig. 2 (a) High-resolution amplitude and phase images for simulation. (b) The output of the CNN based on (a) and different wave vector ($k_{xn}, k_{yn}$)s.

We first analyze the performance using simulation. Figure 2(a) shows the high-resolution object amplitude and phase images. Figure 1(b) shows the CNN output for the low-resolution intensity images with different wave vector ($k_{xn}, k_{yn}$)s. In this simulation, we use 15 by 15 plane waves for illumination and 0.1 NA objective lens to acquire images. The step size for $k_{xn}$ and $k_{yn}$ is 0.05, and the maximal synthetic NA is ~0.64. The pixel size in this simulation is 0.43 μm for the high-resolution object and 1.725 μm for the low-resolution measurements at the object plane (assuming magnification factor is 1). The use of these parameters is to simulate a microscope platform with 2X, 0.1 NA objective and an image sensor with 3.45 μm pixel size.

In Fig. 3, we show the recovered results with different learning rates. The Adam optimizer is used in this simulation. This optimizer is a first-order gradient-based optimizer using adaptive estimates of lower-order moments [37]. It combines the advantages of two extensions of stochastic gradient descent Adaptive Gradient Algorithm (AdaGrad) and Root Mean Square Propagation (RMSProp) [37], and it is the default optimizer for many deep learning problems.

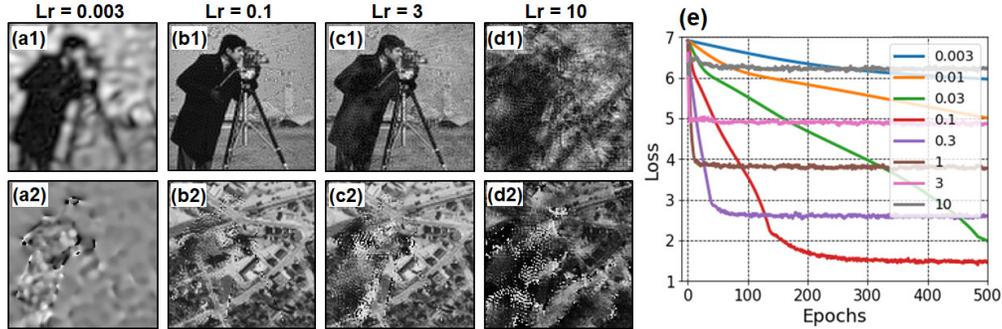

Fig. 3 Different learning rates of the Adam optimizer in TensorFlow for the training process. (a)-(d) The recovered complex object images with learning rates ranging from 0.003 to 10. (e) The L1 loss (in log scale) as a function of epochs. A higher learning rate can decay the loss faster but gets stuck at a worse value of loss. This is because there is too much 'energy' in the optimization process and the learnable weights are bouncing around chaotically, unable to settle in a nice spot in the optimization landscape.

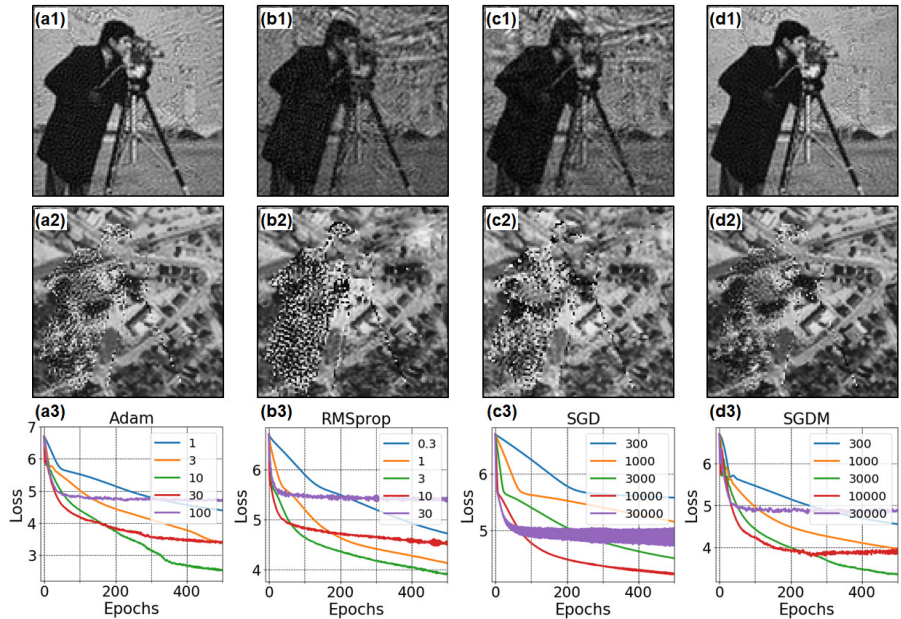

Fig. 4 Performance of different solvers in TensorFlow: (a)Adam, (b) RMSprop, (c) SGD, and (d) SGDM. We use 5 by 5 plane waves for sample illumination and the step size for $k_{xn}$ and $k_{yn}$ is 0.15 in this simulation. The recovered amplitude ((a1)-(d1)) and phase ((a2)-(d2)) with 500 epochs (the best learning rates are chosen in this simulation). Different color curves in (a3)-(d3) represent different learning rates and the loss is in log scale. Adam gives the best performance overall. Batch size is 1 in this simulation.

Different learning rates in Fig. 3 represent different step sizes of the gradient descent approach. We can see that a higher learning rate can decay the loss faster but gets stuck at a worse value of loss. This is because there is too much 'energy' in the optimization process and the learnable weights are bouncing around chaotically, unable to settle in a nice spot in the optimization landscape. On the other hand, a lower learning rate is able to reach a lower minimum point in a slower process. A straight-forward approach for a better learning-rate schedule is to use a large learning rate at the beginning and reduce it for every epoch. How to schedule the learning rate for FP is an interesting topic and requires further investigation in the future.

In Fig. 4, we compare the performance of different optimizers in TensorFlow and show their corresponding recovered results. We note that all optimizers give similar results if the step size for $k_{xn}$ and $k_{yn}$ is small (i.e., aperture overlap is large in the Fourier domain). In this simulation, we use 5 by 5 plane wave illumination with 0.15 step size for $k_{xn}$ and $k_{yn}$. Other parameters are the same as before. Figure 4 shows that Adam achieves the best performance and stochastic gradient descent (SGD) is the worst among the 4. Stochastic gradient descent with momentum (SGDM) is the second-best option. This justifies the adding of momentum in the recent ptychographical iterative engine [38].

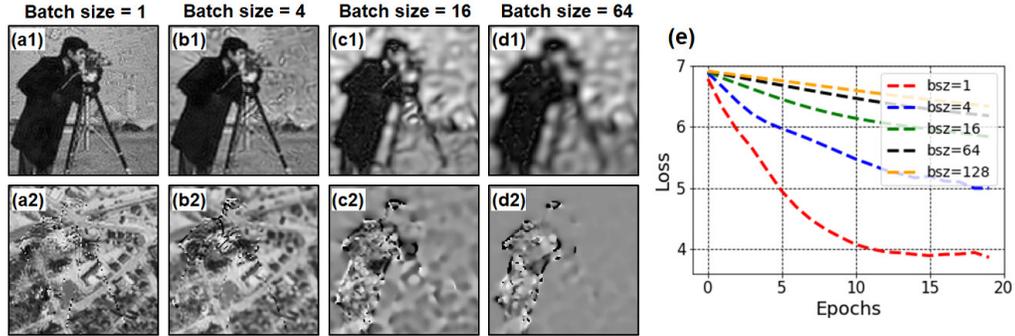

Fig. 5 Performance of different batch sizes as a function of epoch. (a)-(d) The recovered object for different batch sizes and with 20 epochs. (e) The loss (in log scale) with different batch sizes.

In Fig. 5, we investigate the effect of different batch sizes for the optimization process. We can see that batch size of 1 gives the best performance in Fig. 5(a). This justifies the stochastic gradient descent scheme used in the extend ptychography iterative engine (ePIE) [15].

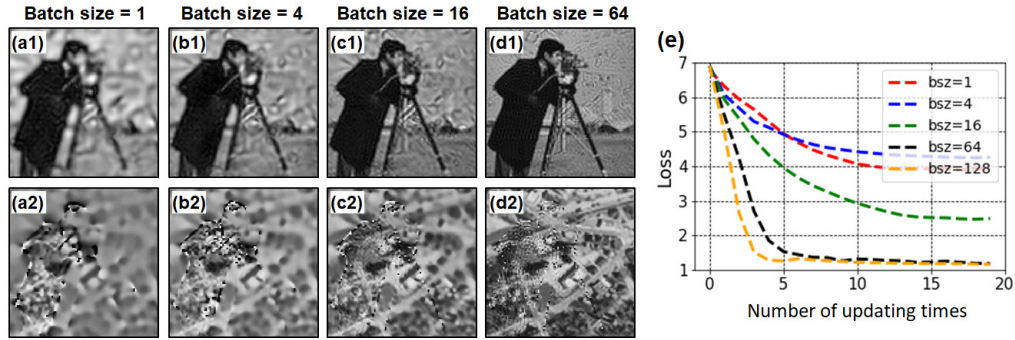

Fig. 6 Performance of different batch sizes as a function of the updating times. (a)-(d) The recovered object for different batch sizes and with 20 updating times. (e) The loss curves (in log scale) with different batch sizes.

However, one advantage of using TensorFlow library is to perform parallel processing via graphics processing unit (GPU) or tensor processing unit (TPU). As a reference point, a modern GPU can handle hundreds of images in one patch. When we use a large batch size, the processing time is about the same as that of batch size = 1. For example, the batch size is 1 in Fig. 5(a) and the epoch number is 20; therefore, we update the object with 225*20 times in this simulation. On the other hand, the batch size is 64 in Fig. 5(d) and the epoch number is 20; therefore, we update the object with (225/64)*20 times for this figure. We define 'number of updating times' as the number of epochs divided by the batch size. This 'number of updating times' is directly related to the processing time of the recovery process.

In Fig. 6(a)-6(d), we show the recovered results with the same number of updating times. In this case, we can see a large batch size leads to a better performance. Based on Figs. 5 and 6, we can draw the following conclusion: batch size of 1 is preferred for serial operation via CPU and a large batch size is preferred for parallel operation using GPU or TPU.

## 3. Modelling Fourier-magnitude-constraint projection in neural network

All widely used iterative phase retrieval algorithms have at their core an operation termed Fourier-magnitude projection (FMP), where an exit complex wave estimate is updated by replacing its Fourier magnitude with measured data while keeping the phase untouched. In the Fourier ptychographic imaging process, the exit complex wave $\widehat{\varphi_n}(k_x, k_y)$ in the Fourier domain can be expressed as

$$\widehat{\varphi_n}(k_x, k_y) = \hat{O}(k_x, k_y) \cdot \text{CTF}_n(k_x, k_y), \tag{7}$$

where $\hat{O}(k_x, k_y)$ and $\text{CTF}_n(k_x, k_y)$ are the Fourier spectrum of the object $O(x, y)$ and the $n^{\text{th}}$ PSF $PSF_n(x, y)$. The FMP operation for the exit complex wave $\widehat{\varphi_n}(k_x, k_y)$ can be written as

$$\widehat{\varphi_n^{update}}(k_x, k_y) = FT\{\sqrt{I_n} \cdot \frac{O(x,y)*PSF_n(x,y)}{|O(x,y)*PSF_n(x,y)|}\} \tag{8}$$

In many ptychographic phase retrieval schemes [15, 21, 33, 38, 39], it is common to accelerate the recovery process by dividing the optimization problem into two sub tasks: 1) perform an FMP to update the exit wave, and 2) the difference between the updated exit wave and the original exit wave is back-propagated to update the object and/or the illumination probe. For Fourier ptychographic imaging process via neural network, we can minimize the following loss function after the FMP in Eq. (8):

$$loss = diff\left(\widehat{\varphi_n^{update}}(k_x, k_y), \hat{O}(k_x, k_y) \cdot \text{CTF}_n(k_x, k_y)\right)$$

$$= \sum_{n=i}^{i-1+batchSize} \left|\widehat{\varphi_n^{update}} - \hat{O} \cdot \text{CTF}_n\right|^2, \tag{9}$$

where L2-norm is used to measure the difference between the updated exit wave and the original exit wave. If the batch size equals to 1, Eq. (9) is similar the ePIE scheme [15].

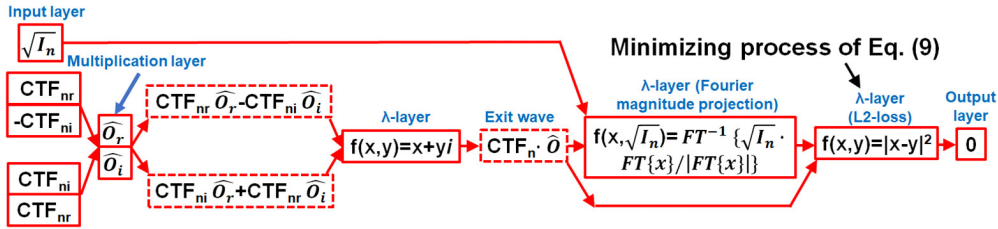

Fig. 7 The multiplication neural network model for the Fourier ptychographic imaging process in Eq. (9). The input of the network is the $n^{\text{th}}$ CTF and the measured data $\sqrt{I_n}$. The object's Fourier spectrum is modeled as learnable weights of a multiplication layer. We use a λ-layer to define the Fourier-magnitude-projection operation. The output of the network is 0, and thus, the training process of the network minimizes the loss function defined in Eq. (9). The implementation code is provided in dataset 2.

The neural network for minimizing Eq. (9) is shown in Fig. 7, where we model the object's Fourier spectrum as trainable weights of a multiplication layer. Since the trainable weights need to be real in TensorFlow implementation, we separate the complex object spectrum $\hat{O}(k_x, k_y)$ into two channels with subscripts 'r' and 'i' representing the real and imaginary parts in Fig. 7. The input layer for this network is the $n^{\text{th}}$ $\text{CTF}_n$ and the captured low-resolution amplitude $\sqrt{I_n}$. We use a λ-layer in TensorFlow to define the Fourier-magnitude-projection operation. The output of the network is 0, and thus, the training process of the network minimizes the loss function defined in Eq. (9). Once the complex object spectrum $\hat{O}(k_x, k_y)$ is recovered in the network training process, we can perform an inverse Fourier transform to get the complex object $O(x, y)$ in the spatial domain.

In Fig. 8, we compare the recovered results in 4 cases: 1) minimizing the loss function in Eq. (6) with L2-norm ('L2-norm intensity' in Fig 8(a)), 2) minimizing the loss function in Eq. (6) with L1-norm ('L1-norm intensity' in Fig. 8(b)), 3) minimizing the loss function in Eq. (9) with L2-norm ('L2-norm exit wave' in Fig. 8(c)), and 4) minimizing the loss function in Eq. (9) with L1-norm ('L1-norm exit wave' in Fig. 8(d)). We can see that the cases of 'L1-norm intensity' and 'L2-norm exit wave' give the best results. We also note that the intensity updating cases tend to recover the low-resolution features first while the exit-wave updating cases tend to recover features at all levels. This behavior can be explained by the loss functions in Eqs. (6) and (9). The loss function in Eq. (6) is to reduce the difference between two intensity images in the spatial domain. Therefore, it tends to correct the low-frequency difference first since most energy concentrates in this region. On the other hand, the loss function in Eq. (9) is to reduce the difference between two Fourier spectrums and it does not focus on the low-frequency regions. As such, the resolution improvement is more obvious for the exit-wave updating cases.

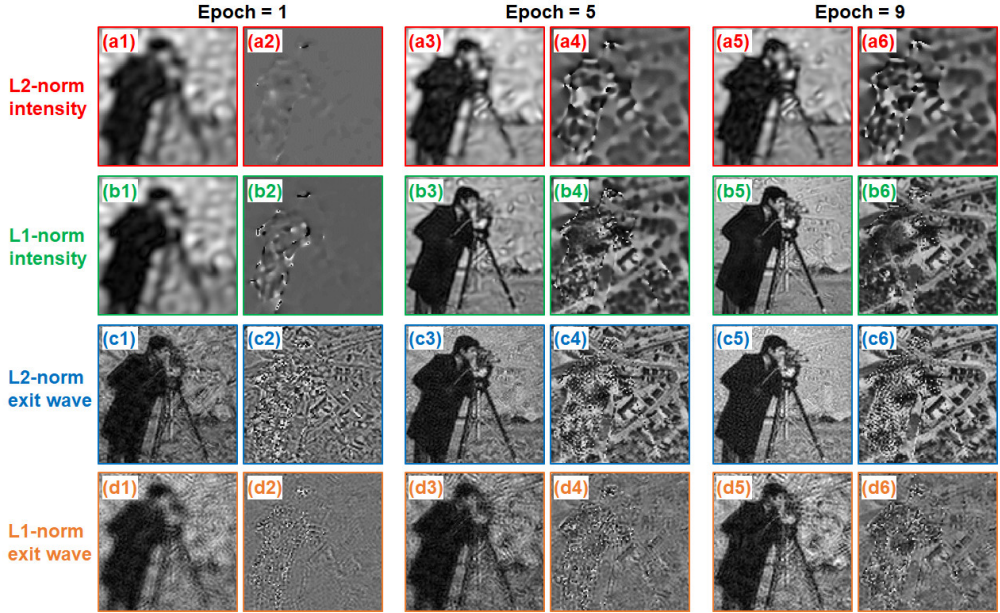

Fig. 8 Comparison of different cases with batch size = 1. The learning rates are chosen based on the fastest loss decay in 10 epochs. (a) Minimizing the loss function in Eq. (6) with L2-norm, termed 'L2-norm intensity'. (b) Minimizing the loss function in Eq. (6) with L1-norm, termed 'L1-norm intensity'. (c) Minimizing the loss function in Eq. (9) with L2-norm, termed 'L2-norm exit wave'. (d) Minimizing the loss function in Eq. (9) with L1-norm, termed 'L1-norm exit wave'. The resolution improvement is more obvious for the exit-wave cases.

In Fig. 9, we have tested the L2-norm exit-wave network using experimental data. Figure 9(a) shows the experimental setup where we use a 2X, 0.1 NA Nikon objective with a 200 mm tube lens (Thorlabs TTL200) to build a microscope platform. A 5-megapixel camera (BFS-U3-51S5M-C) with a 3.45 μm pixel size is used to acquire the intensity images. We use an LED array (Adafruit 32 by 32 RGB LED matrix) to illuminate the sample from different incident angles and the distance between the LED array and the sample is ~85 mm. In this experiment, we illuminate the sample from 15 by 15 different incident angles and the corresponding maximum synthetic NA is ~0.55. Figure 9(b1)-9(d1) show the low-resolution images captured by the microscope platform in Fig. 9(a). We use the L2-norm exit-wave network with the Adam optimizer to recover the complex object spectrum in the multiplication layer. The batch size in this experiment is 1 and we use 20 epochs in the network training (optimization) process. The recovered object intensity images are shown in Fig. 9(b2)-9(d2) and the recovered phase images are shown in Fig. 9(b3)-9(d3). In Fig. 9(b4)-(d4), we combine three recovered intensity images captured using red, green, and blue LEDs. We can clearly see the resolution improvement from the network reconstruction.

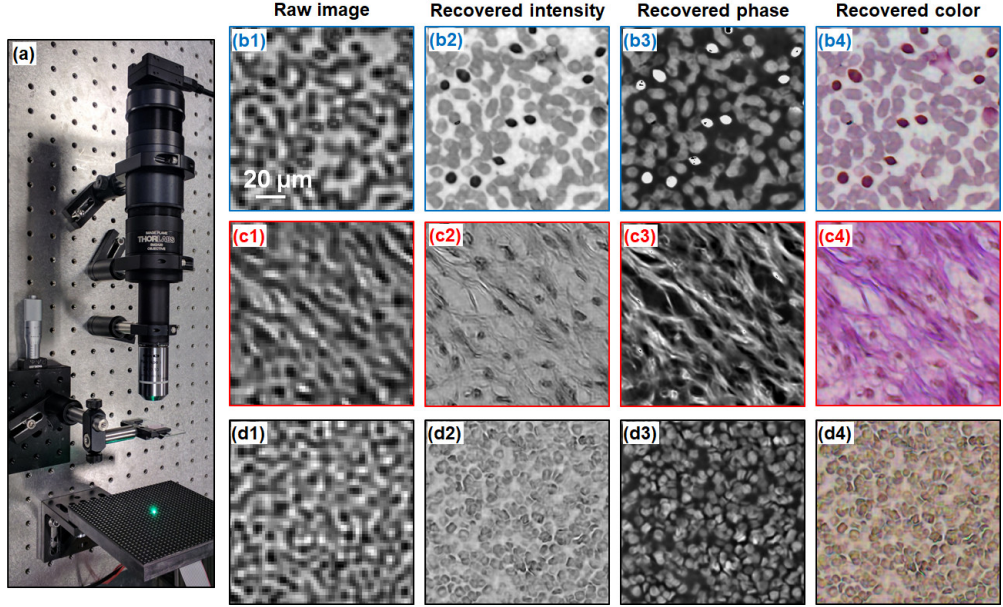

Fig. 9 Test the L2-norm exit-wave network with experimental data. We use Adam optimizer with 20 epochs in this experiment and the batch size is 1. (a) The experimental setup with a 2X, 0.1 NA objective lens and a 3.45 μm pixel size camera, which is the same as the simulation setting. We test three different samples: (b) a blood smear, (c) a brain slide, and (d) a tissue section stained by immunohistochemistry methodology. (b1)-(d1) show the captured raw images using the 2X objective lens. The recovered intensity images are shown in (b2)-(d2) and the recovered phase images are shown in (b3)-(d3). (b4)-(d4) show the recovered color images by combining three recovered intensity images at 632 nm, 532 nm, and 470 nm wavelengths.

## 4.  Extensions for single-pixel imaging and structured illumination microscopy

Here we extend the network models discussed above for single-pixel imaging and structured illumination microscopy. Single pixel imaging captures images using single-pixel detectors. It enables imaging in a variety of situations that are impossible or challenging with conventional 2D image sensors [40-42]. The forward imaging process of single-pixel imaging can be expressed as

$$I_n = \sum_{x,y} O(x,y) \cdot P_n(x,y), \qquad (10)$$

where $O(x,y)$ denotes the 2D object, $P_n(x,y)$ denotes the $n^{th}$ 2D illumination pattern on the object, and $I_n$ denotes the $n^{th}$ single-pixel measurement. The summation sign in Eq. (10) represents the signal summation over the x-y plane. Since the dimensions of the object and pattern are the same, the forward imaging model in Eq. (10) can be modeled by a 'valid convolution layer' which outputs a predicted single-pixel measurement. The CNN model for single-pixel imaging is shown in Fig. 10(a). The training of this model is to minizine the following loss function:

$$loss = diff\left(I_n, \sum_{x,y} O(x,y) \cdot P_n(x,y)\right) = \sum_{n=i}^{i-1+batchSize}\left|I_n - \sum_{x,y} O(x,y) \cdot P_n(x,y)\right|, \quad (11)$$

where L1-norm is used to measure the difference between the network prediction and the actual measurements. A detailed analysis of the different solvers' performance and different regularization schemes is beyond the scope this paper and it will be presented elsewhere.

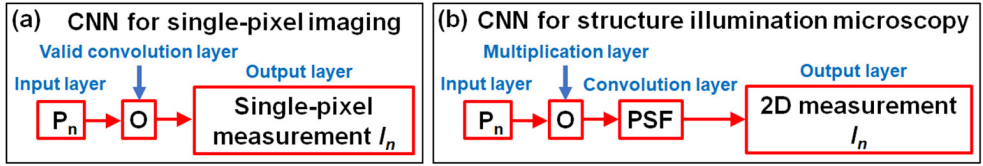

Fig. 10 CNN models for (a) single-pixel imaging and (b) structured illumination microscopy. For single-pixel imaging, the input is the illumination pattern $P_n(x,y)$, the object is modeled as learnable weights in a valid convolution layer, and the output is the predicted single-pixel measurement. Similarly, for structured illumination microscopy, the object is modeled as learnable weights for a multiplication layer and the output is the predicted 2D image.

Structured illumination microscopy (SIM) uses non-uniform patterns for sample illumination and combine multiple acquisitions for super-resolution image recovery [43-45]. Frequency mixing between the sample and the non-uniform illumination pattern modulates the high-frequency components to the passband of the collection optics. Therefore, the recorded images contain sample information that is beyond the limit of the employed optics. Conventional SIM employs sinusoidal patterns for sample illumination. In a typical implementation, three different lateral phase shifts (0, $2\pi/3$, $4\pi/3$) are needed for each orientation of the sinusoidal pattern, and 3 different orientations are needed to double the bandwidth isotropically in the Fourier domain. Therefore, 9 acquisitions are needed.

The forward imaging process of SIM can be expressed as

$$I_n(x,y) = \left(O(x,y) \cdot P_n(x,y)\right) * PSF, \quad (12)$$

where $O(x,y)$ denotes the 2D object, $P_n(x,y)$ denotes the $n^{th}$ 2D illumination pattern on the object, PSF denotes the PSF of the objective lens, and $I_n(x,y)$ denotes the $n^{th}$ 2D image measurement. The CNN model for SIM is shown in Fig. 10(b), where the object is modeled as learning weights of a multiplication layer. The training of this model is to minizine the following loss function:

$$loss = diff\left(I_n, \left(O(x,y) \cdot P_n(x,y)\right) * PSF\right)$$
$$= \sum_{n=i}^{i-1+batchSize}\left|I_n - \left(O(x,y) \cdot P_n(x,y)\right) * PSF\right|, \quad (13)$$

where L1-norm is used to measure the difference between the network prediction and the actual measurements. Figure 11 shows our simulation of using the proposed CNN model for SIM reconstruction. Figure 11(a) shows the object image under uniform illumination; the resolution of this image represents the resolution of the employed objective lens. Different from the conventional implementation using 9 illumination patterns, we use 4 sinusoidal patterns for sample illumination and the 4 SIM measurements are shown in Fig. 11(b). The recovered super-resolution image using the proposed CNN is shown in Fig. 11(c). The minimization process via Eq. (13) is able to reduce the number of acquisitions for SIM.

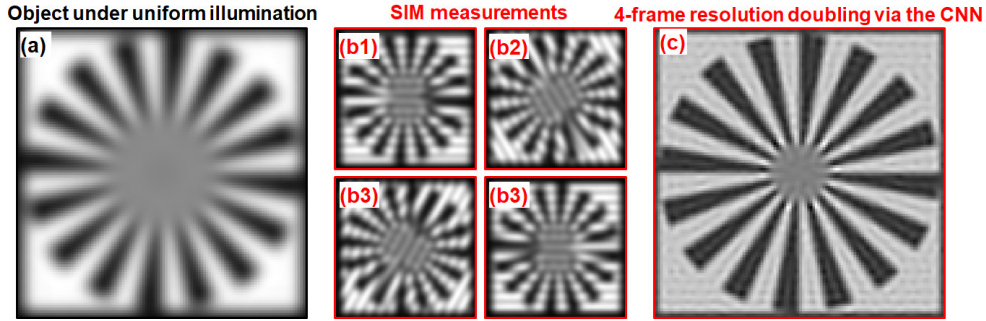

Fig. 11 4-frame resolution doubling using the CNN-based SIM model. (a) The object image under uniform illumination. (b) The 4 SIM measurements using 4 sinusoidal patterns for sample illumination. The resolution doubled image recovered by the training process of the CNN in Fig. 10(b).

## 5. Summary and discussion

In summary, we model the Fourier ptychographic forward imaging process using a convolution neural network (CNN) and recover the complex object information in the network training process. In our approach, the object is treated as 2D learnable weights of a convolution or a multiplication layer. The output of the network is modeled as the loss function we aim to minimize. The batch size of the network corresponds to the number of captured low-resolution images in one forward / backward pass. We use the popular open-source machine learning library, TensorFlow, for setting up the network and conducting the optimization process. We show that the Adam optimizer achieves the best performance in general and a large batch size is preferred for GPU / TPU-based parallel processing.

Another contribution of our work is to model the Fourier-magnitude projection via a neural network model. The Fourier-magnitude projection is the most important operation in iterative phase retrieval algorithms. Based on our model, we can easily perform exit-wave-based optimization using TensorFlow. We show that L2-norm is preferred for exit-wave-based optimization while L1-norm is preferred for intensity-based optimization.

Since convolution and multiplication are the two most-common operations in imaging modeling, the reported approach may provide a new perspective to examine many coherent and incoherent systems. As a demonstration, we discuss the extensions of the reported networks for modeling single-pixel imaging and structured illumination microscopy. We show that single-pixel imaging can be modeled by a convolution layer implementing 'valid convolution'. For structured illumination microscopy, we propose a network model with one multiplication layer and one convolution layer. In particular, we demonstrate 4-frame resolution doubling via the proposed CNN. Since the proposed network model can be implemented in neural engine and TPU, we envision many opportunities for accelerating the image reconstruction process via machine-learning hardware.

There are many future directions for this work and our efforts are on-going. First, we can implement the CTF updating scheme in the proposed neural network models. One solution is to make the incident wave vector as the input and we can then convolute the CTF with $\delta(k_{xn}, k_{yn})$ to generate $CTF_n$. In this case, we can model CTF as learnable weights in a convolution layer and it can be updated in the network training process. Second, correcting positional error is an important topic for real-space ptychographic imaging. The positional errors in real-space ptychography is equivalent to the errors of incident wave vectors in FP. We can, for example, model $(k_{xn}, k_{yn})$ as learnable weights in a layer and they can be updated in the network training process. Similarly, we can also generate CTF based on coefficients of different Zernike modes and model such coefficients as learnable weights. Third, the proposed network models are developed for one coherent state. It is straight forward to extend our networks to model multi-state cases. Fourth, we use a fixed learning rate in our models. How to schedule the learning rates for faster convergence is an interesting topic and requires further investigations. Fifth, we can add regularization term such as total variation loss in the model to better handle the measurement noises.

We provide our implantation code in [datasets 1-2](#) in the format of Jupyter notebook.